\documentclass{article} %
\usepackage{latexml}
\usepackage[top=2.25cm,bottom=2.5cm,left=2.5cm,right=2.5cm,columnsep=0.65cm]{geometry}
\usepackage[T1]{fontenc}
\usepackage[scaled=0.92]{helvet}
\usepackage{microtype,parskip,graphicx,xcolor}
\usepackage[round,authoryear]{natbib}
\usepackage{titlesec,caption,tcolorbox}
\tcbuselibrary{skins,breakable}
\usepackage[hyperfootnotes=false]{hyperref}

\definecolor{PaperPanel}{HTML}{F1F3F5}
\definecolor{PaperBorder}{HTML}{D8DCE2}
\definecolor{PaperInk}{HTML}{202124}
\definecolor{PaperLink}{HTML}{42474D}
\definecolor{PaperResourceLink}{cmyk}{0,1,0,0}
\hypersetup{colorlinks=true,linkcolor=PaperLink,citecolor=PaperLink,urlcolor=PaperLink}
\setcitestyle{authoryear,round,citesep={;},aysep={,},yysep={;}}
\renewcommand{\bibfont}{\small}

\titleformat*{\section}{\Large\sffamily\bfseries}
\titleformat*{\subsection}{\large\sffamily\bfseries}
\titleformat*{\subsubsection}{\normalsize\sffamily\bfseries}
\titleformat*{\paragraph}{\normalsize\sffamily\bfseries}
\DeclareTextFontCommand{\textbf}{\bfseries\sffamily}
\makeatletter
\newenvironment{frontmatter}{%
  \begingroup
  \iflatexml
  \else
  \begin{tcolorbox}[
    enhanced,frame hidden,boxrule=0pt,arc=8pt,
    colback=PaperPanel,coltext=PaperInk,
    left=14pt,right=14pt,top=13pt,bottom=13pt,
    before skip=0pt,after skip=12pt
  ]
  \fi
}{%
  \iflatexml
    \par
  \else
  \end{tcolorbox}%
  \fi
  \@thanks
  \endgroup
  \setcounter{footnote}{0}%
  \gdef\@thanks{}%
}
\renewcommand{\maketitle}{%
  {\raggedright\sffamily\bfseries\fontsize{20}{24}\selectfont\@title\par}%
  \nobreak\vspace{6pt}%
  {\raggedright\sffamily\bfseries\normalsize\@author\par}%
  \vspace{10pt}%
}
\renewenvironment{abstract}{\par\noindent\ignorespaces}{\par}
\makeatother

\usepackage{amsmath,amsfonts,bm}

\def\eqref#1{equation~\ref{#1}}

\def\1{\bm{1}}

\DeclareMathAlphabet{\mathsfit}{\encodingdefault}{\sfdefault}{m}{sl}
\SetMathAlphabet{\mathsfit}{bold}{\encodingdefault}{\sfdefault}{bx}{n}

\usepackage{hyperref}
\usepackage{url}

\usepackage{multirow}  %
\usepackage{algorithm}
\usepackage{xcolor}
\usepackage{amsmath}
\usepackage{amssymb}
\usepackage{algpseudocode}
\usepackage{subcaption}
\usepackage{tikz}
\usetikzlibrary{shadings}

\usepackage{xcolor}

\usepackage[utf8]{inputenc} %
\usepackage[T1]{fontenc}    %
\usepackage{hyperref}       %
\usepackage{url}            %
\usepackage{booktabs}       %
\usepackage{amsfonts}       %
\usepackage{nicefrac}       %
\usepackage{microtype}      %
\usepackage{xcolor}         %
\usepackage{amsmath}
\usepackage{multirow}
\usepackage{multicol}
\usepackage{colortbl}
\usepackage{tcolorbox}
\usepackage{makecell}
\usepackage{wrapfig}
\definecolor{light-gray}{gray}{0.6}
\definecolor{front-color}{HTML}{F5FFFA}
\definecolor{Gray}{gray}{0.93}
\definecolor{myblue}{rgb}{0.784, 0.863, 0.961}
\definecolor{myyellow}{rgb}{0.980, 0.961, 0.863}
\usepackage{multirow} 
\usepackage{booktabs}

\usepackage{xcolor}

\definecolor{PromptBorder}{HTML}{065F5C}  %
\definecolor{PromptBack}{HTML}{F2F9F8}      %

\let\PromptOriginalTT\ttdefault
\usepackage{inconsolata}
\let\ttdefault\PromptOriginalTT
\tcbuselibrary{listings,skins,breakable}

\newtcblisting{promptbox}[1]{
    breakable,                 %
    colframe=PromptBorder,     %
    colback=PromptBack,        %
    coltitle=white,            %
    fonttitle=\bfseries,       %
    title={#1},                %
    arc=3pt,                   %
    boxrule=1.5pt,             %
    left=6pt, right=6pt, top=4pt, bottom=4pt, %
    listing only,
    listing options={
  basicstyle=\fontencoding{T1}\fontfamily{zi4}\fontsize{9}{9}\selectfont,
  breaklines=true,
  breakatwhitespace=true,
  columns=fullflexible,
  keepspaces=true,
  showstringspaces=false,
  escapeinside={(*@}{@*)}
}
}

\definecolor{TealBorder}{HTML}{065F5C} 
\definecolor{TealBack}{HTML}{F2F9F8}   

\newtcolorbox{skillbox}[1]{
    breakable,
    colframe=TealBorder,
    colback=TealBack,
    coltitle=white,
    fonttitle=\bfseries,
    title={#1},
    arc=3pt,
    boxrule=1.5pt,
    left=6pt, right=6pt, top=4pt, bottom=4pt
}

\definecolor{caseRed}{RGB}{255,190,190}
\definecolor{caseGreen}{RGB}{0,255,60}
\definecolor{caseTeal}{RGB}{0,210,210}
\definecolor{casePurple}{RGB}{215,170,230}
\definecolor{caseYellow}{RGB}{255,245,0}

\usepackage{pifont}
\newcommand{\cmark}{\textcolor{green!60!black}{\ding{51}}}
\newcommand{\xmark}{\textcolor{red!70!black}{\ding{55}}}
\usepackage{xcolor}

\usepackage{enumitem}
\title{VideoMemBench: How Well Do MLLMs Remember\\in Video Conversations?}

\input{authors.tex}

\newcommand{\eg}{\emph{e.g}.}

\hypersetup{pdftitle={VideoMemBench: How Well Do MLLMs Remember in Video Conversations?}}
\begin{document}

\begin{frontmatter}
\maketitle

\begin{abstract}
Multimodal large language models (MLLMs) can now process hour-long videos, making extended multi-turn dialogue over a single video a natural mode of use. 
Sustaining such a dialogue requires more than answering each question in isolation: the model must retrieve and reuse information established earlier in the conversation, both what it has seen and what it has said, an ability we term {in-context video memory}. 
Existing video QA benchmarks leave this ability unmeasured: their questions are mutually independent, never reference one another, and each is answerable from the video alone, so nothing ever requires carrying information across turns. 
We introduce \textbf{VideoMemBench}, in which each sample pairs a single video with a long multi-turn dialogue, whose turns are connected by typed cross-turn dependencies: an {Episodic Dependency} refers back to a visual episode described in an earlier turn, probing memory of {what the model has seen}, while an {Answer Dependency} builds on the model's own earlier answer, probing memory of {what the model has said}. 
Every dependency is verified considering two properties: {1) necessity}, meaning the question cannot be answered without its earlier turns, and {2) non-leakage}, meaning the question does not restate the information from its earlier turns. 
To enforce these properties, we design a construction pipeline with three LLM roles: a planner, a generator, and a judger, with each dialogue reviewed by human experts.
VideoMemBench comprises 326 videos with 2,476 turns, of which 1,387 depend on earlier turns, including multi-parent dependencies and long-range call-back.
Among over 20 evaluated open- and closed-source MLLMs, Claude Opus 5 performs the best with an overall score of only 45.7\%, highlighting the challenge of in-context video memory.
Our dataset and evaluation are released.

\vspace{0.8em}
{\small
\textbf{Project:} \href{https://videomembench.github.io}{\textcolor{PaperResourceLink}{https://videomembench.github.io}}\\[2pt]
\textbf{Video:} \href{https://huggingface.co/datasets/VideoMemBench/data}{\textcolor{PaperResourceLink}{https://huggingface.co/datasets/VideoMemBench/data}}\\[2pt]
\textbf{GitHub:} \href{https://github.com/Oliiveralien/VideoMemBench}{\textcolor{PaperResourceLink}{https://github.com/Oliiveralien/VideoMemBench}}
}
\end{abstract}
\end{frontmatter}

\section{Introduction}

Frontier multimodal large language models (MLLMs)~\citep{Gemini,GPT5,Kimi,Qwen3,Claude} can now process hour-long videos, with million-token contexts that accommodate an entire film, a full match, or a day-long recording. 
Building on this capability, extended multi-turn dialogue over a single long video is emerging as a natural mode of use: the model accompanies a user through the video with a stream of follow-up questions about its characters, events, and details. 
Sustaining such a conversation raises a key question: when consecutive questions concern the same video, can the model correctly retrieve and reuse what the dialogue has already established? We refer to this ability as \emph{in-context video memory}.

Recently, in-context memory has been systematically evaluated for large language models (LLMs) in text-only settings~\citep{maharana2024locomo, hsieh2024ruler,kuratov2024babilong,tan2025membench,wu2025longmemeval,huet2025episodic}. 
However, extending this evaluation to video poses a key challenge: distinguishing questions that genuinely depend on prior dialogue context from those answerable using only the video and the current question.
Existing video understanding benchmarks~\citep{videomme, mvbench, egoschema, longvideobench, lvbench, mmev2} are structurally unable to evaluate it. 
Typically, they decompose the QA evaluation into mutually independent questions: no question references the content or answer of another, so nothing ever needs to be carried across turns, and every question is answerable from the video alone.
In other words, their dialogue histories are never load-bearing.

\begin{figure*}[t]
\centering
\includegraphics[width=\linewidth]{figs/teaser.pdf}
\caption{\textbf{Example of VideoMemBench with cross-turn dependencies:}
Independent turns (\eg, $T_1$, $T_8$) establish visually grounded facts;
dependent turns refer back to earlier turns through expressions
such as ``the scene from $T_1$'' or ``the target from $T_1$'', without restating earlier content.
An \emph{Episodic Dependency} (\eg, $T_2$) requires recalling \emph{what the model has seen}: the model must recover which scene $T_1$ described before answering the question. 
An \emph{Answer Dependency} (\eg, $T_3$, $T_4$) requires recalling \emph{what the model has said}: each builds on the answer produced at an earlier turn. 
}
\label{fig:teaser}
\end{figure*}

To this end, we introduce \textbf{VideoMemBench}. 
Each sample pairs a single video with a long multi-turn dialogue whose turns are connected by explicit, \emph{typed} cross-turn dependencies, so that a sample is not a concatenation of separate questions but a memory chain running through the entire conversation. 
The key design is that answering a dependent turn should require the dialogue and the video jointly: the dialogue supplies what an earlier turn established, and the video supplies the visual facts to ground the current one, so that neither source alone suffices. 
Specifically, we design two dependency types, enforcing two forms of memory: \textit{Episodic Dependency}, for recalling previously discussed visual episodes, and \textit{Answer Dependency}, for recalling the model's own earlier responses.

\textbf{(1) Episodic Dependency} arises when a question depends on the visual episode that an earlier turn described: only by recalling that referent from the dialogue can the model determine what is being asked. 
As shown in \autoref{fig:teaser}, $T_1$ asks about the suspect mugshot pinned to the briefing-room corkboard, and $T_2$ then asks whether the hippo officers are present in ``the scene from $T_1$'' and what their names are, without restating where that scene is. 
Unless the model retains that $T_1$ was about the briefing room, it cannot tell which part of the video ``the scene from $T_1$'' refers to, let alone check who is in it. 
Episodic dependencies thus hinge on memory of \emph{what the model has seen}.

\textbf{(2) Answer Dependency} arises when a question builds on the answer the model itself gave at an earlier turn: the model must continue from its own earlier response. 
As shown in \autoref{fig:teaser}, $T_1$ establishes that the suspect is an anteater, $T_3$ then asks whether the animal that fox and rabbit meet at the shipping-container yard is ``the target from $T_1$'', which cannot be decided without the answer given at $T_1$. 
$T_4$ in turn asks what happens next to ``the animal met in $T_3$'', building on the answer to $T_3$. 
Answering thus demands two abilities at once: \emph{retrieval}, locating the earlier answer within the accumulated dialogue, and \emph{consistency}, keeping the current answer faithful to it rather than silently re-judging the video. 
Answer dependencies thus hinge on memory of \emph{what the model has said}.

Furthermore, as shown in the closing $T_9$ of \autoref{fig:teaser}, VideoMemBench stresses memory along two additional dimensions: 
a turn may depend on several earlier turns at once, requiring parallel recall of multiple episodes ($T_1$) or answers ($T_8$); and a dependency may span eight turns ($T_1 \rightarrow T_9$), keeping content from the beginning of the dialogue live throughout. 

\begin{table*}[t]
\centering
\caption{\textbf{Comparison with existing benchmarks.}
\emph{Max Dur.} is the maximum video duration in the benchmark.
\emph{Multi-turn} indicates multi-turn dialogue evaluation.
\emph{Max Turns} is the largest number of QA turns in one dialogue.
\emph{Dep. Graph} denotes structured dependency graphs for individual turns; 
\emph{Typed} distinguishes answer and episodic dependencies; \emph{M-Parent} indicates multiple parents.
}
\label{tab:compare}
\small
\setlength{\tabcolsep}{3.2pt}
\renewcommand{\arraystretch}{1.00}
\resizebox{\textwidth}{!}{%
\begin{tabular}{llrrrrccccc}
\toprule
\makecell[cl]{Benchmark} & \makecell{Input} & \makecell{\#Videos} & \makecell{\#Dialogs} & \makecell{\#QAs} & \makecell{Max Dur.} & \makecell{Max Turns} & \makecell{Multi-turn} & \makecell{Dep.\\Graph} & \makecell{Typed} & \makecell{M-Parent} \\
\midrule
MVBench~\citep{mvbench}
  & Video & 3,641 & -- & 4,000 & 3\,min & 1 & \xmark & \xmark & \xmark & \xmark \\
Video-MME~\citep{videomme}
  & Video & 900 & -- & 2,700 & 1\,h & 1 & \xmark & \xmark & \xmark & \xmark \\
LVBench~\citep{lvbench}
  & Video & 103 & -- & 1,549 & 2\,h & 1 & \xmark & \xmark & \xmark & \xmark \\
\midrule
MMDU~\citep{mmdu}
  & Images & -- & 110 & 1,645 & -- & 27 & \cmark & \xmark & \xmark & \xmark \\
MT-Video-Bench~\citep{Mt-video-bench}
  & Video & 135 & 1,000 & 5,887 & 20\,min & 8 & \cmark & \xmark & \xmark & \xmark \\
\midrule
\textbf{VideoMemBench (ours)}
  & \textbf{Video} & \textbf{326} & \textbf{326} & \textbf{2,476} & \textbf{23.47\,h} & \textbf{21} & \cmark & \cmark & \cmark & \cmark \\
\bottomrule
\end{tabular}%
}
\end{table*}

These structurally diverse dependencies ensure that answering questions from our \textbf{VideoMemBench} requires the model to exercise \emph{in-context video memory} throughout the dialogue.
We further verify every dependent turn for \emph{necessity} and \emph{non-leakage}: the question must be unanswerable without its parent turns, and must not restate the information they established before. 
In total, VideoMemBench comprises 326 videos, each paired with a human-reviewed dialogue, totaling 2,476 turns, of which 1,387 depend on earlier turns.
Videos are drawn from five existing benchmarks and the web, covering 31 semantic domains across four categories (lifestyle \& entertainment, knowledge \& education, art \& literature, sports \& competition), and last up to 23.47 hours each. 
The comparison in \autoref{tab:compare} shows that single-turn benchmarks evaluate questions independently, whereas multi-turn benchmarks extend evaluation to dialogues. VideoMemBench further provides explicit dependency graphs with typed edges and multiple parents, enabling structured evaluation of in-context video memory.

Evaluating over 20 representative open- and closed-source MLLMs on VideoMemBench, we find that current models clearly struggle with cross-turn dependencies: the best model we tested (\textit{Claude Opus 5}) achieves an overall score of only 45.7\%, with 35.7\% on answer-dependent turns and 49.1\% on episodic-dependent turns, both below 53.9\% on independent turns.
These results indicate that in-context video memory remains an open problem for current MLLMs.

Our main contributions are summarized as follows:
\begin{itemize}[leftmargin=*]
\item We study \emph{in-context video memory}, the ability of a model to retrieve and reuse what earlier turns have established in a consecutive video QA dialogue, which existing benchmarks leave unmeasured.

\item We build \textbf{VideoMemBench}, in which each sample is a long multi-turn dialogue grounded in a single video, with turns connected by typed cross-turn \textit{Episodic} and \textit{Answer} dependencies. 
VideoMemBench comprises 326 video dialogues, totaling 2,476 turns with 1,387 dependent, verified for necessity and non-leakage, and reviewed by human experts.

\item  We conduct extensive evaluations of over 20 open- and closed-source MLLMs on VideoMemBench. The best model achieves only 45.7\% overall, with performance declining as dependency complexity increases, indicating that in-context video memory remains challenging for current MLLMs.

\end{itemize}

\section{Related Works}

\textbf{Video Understanding Benchmarks.} 
Existing video understanding benchmarks usually focus on temporal perception and reasoning.
MVBench~\citep{mvbench} and TempCompass~\citep{tempcompass} evaluate temporal understanding, while EgoSchema~\citep{egoschema} targets long-form video QA.
Broadening the scope, Video-MME~\citep{videomme} covers diverse video durations and tasks, and Video-MMMU~\citep{videommmu} evaluates knowledge acquisition from multidisciplinary videos.
For longer contexts, LongVideoBench~\citep{longvideobench} and LVBench~\citep{lvbench} focus on long-video understanding, while EgoLife~\citep{egolife} examines egocentric experiences spanning hours and days.
Beyond temporal understanding, VSI-Bench~\citep{vsibench} evaluates video-based spatial reasoning.
Despite this growing coverage and temporal range, these benchmarks evaluate questions independently, without requiring information established in the earlier dialogue.

\textbf{Memory and Multi-Turn Evaluation.} 
Existing benchmarks evaluate conversational memory and multi-turn interaction from several perspectives. 
LoCoMo~\citep{maharana2024locomo} and LongMemEval~\citep{wu2025longmemeval} assess long-term conversational memory, while MemBench~\citep{tan2025membench} distinguishes factual and reflective memory. 
MT-Bench~\citep{mtbench} evaluates two-turn response quality. 
In multimodal settings, MMDU~\citep{mmdu} evaluates multi-turn, multi-image conversations, and MMRC~\citep{mmrc} includes multi-turn image reasoning and memory recall, but neither targets dialogue memory grounded in a single video. 
VideoMemBench addresses this gap through explicit episodic and answer dependencies that link later questions to previously discussed video scenes and generated answers, with support for multiple parents and long-range connections.

\begin{figure*}[t]
\centering
\includegraphics[width=1\linewidth]{figs/construction_pipeline.pdf}
\caption{\textbf{The VideoMemBench construction pipeline.}
Starting from a curated video pool, a planner specifies turn intents and a typed dependency graph. A generator constructs the dialogue turn by turn, while a judger checks necessity, non-leakage, and visual grounding, returning failed turns for regeneration. Completed dialogues are checked against the plan and reviewed by human experts.}
\label{fig:construction_pipeline}
\end{figure*}

\section{VideoMemBench}
\label{sec:bench}

Each VideoMemBench sample pairs a video with a multi-turn dialogue, annotated with reference answers and a typed dependency graph. 
Each edge links a dependent turn to a parent turn and is labeled as Episodic or Answer. 
Below, we describe the requirements for valid dependencies and visually grounded answers, followed by the construction pipeline and dataset statistics.

\subsection{Design Requirements}
\label{sec:requirements}

Constructing valid dependencies requires going beyond `simple references to earlier turns.
The key challenge lies in ensuring that cross-turn dependencies are genuinely necessary, do not leak prior information, and remain grounded in the video.
This leads to three requirements:

\noindent \textbf{Req-1: Dependency Necessity.} 
A dependent question must become unanswerable when its parent turns are removed, even with access to the video and the remaining dialogue. 
Considering a cooking video, an implicit reference alone does not ensure this: asking how much of \texttt{``that ingredient''} is used requires no dialogue history if the video contains only one ingredient. Such a question avoids revealing the ingredient's identity but fails the necessity requirement.

\noindent \textbf{Req-2: Dependency Non-leakage.} 
A dependent question must not restate the episode descriptions or answers from its parents, even partially. 
For example,  if an earlier answer identifies \texttt{``the flour in the red bowl''}, asking \texttt{``How much of the flour identified earlier is used?''} partially reveals that answer. 
If both a red and a blue bowl contain flour, the parent remains necessary to identify the target bowl. 
Thus, a turn may satisfy necessity while violating non-leakage.

\noindent \textbf{Req-3: Visual grounding.} 
Every turn, whether dependent or not, must be anchored in a verifiable visual fact in the video.
For dependent turns, this evidence must support the answer under the interpretation established by the parent turns. 

\subsection{Construction Pipeline}
\label{sec:pipeline}

Guided by these requirements, we construct VideoMemBench through a pipeline combining three LLM roles with human expert review, as shown in \autoref{fig:construction_pipeline}.
A \emph{planner} lays out the dependency structure of a dialogue before any question is written, providing the structural precondition for necessity (Req-1). 
A \emph{generator} realizes the plan turn by turn, referring to earlier turns only through referring expressions (Req-2) and anchoring every answer in frame evidence (Req-3). 
A \emph{judger} then verifies each generated turn against the three requirements and checks completed dialogues for consistency with the initial structural plan.
Human experts finally review each retained dialogue. 

\noindent$\bullet$ \textbf{Video Collection.}
We collect videos from multiple sources: Video-MME~\citep{videomme}, VSI-Bench~\citep{vsibench}, Video-MMMU~\citep{videommmu}, LongVideoBench~\citep{longvideobench} and the web. 
Web videos are sourced from YouTube across sports, talks and lectures, tutorials, film recaps, and cooking.
We also include EgoLife~\citep{egolife} recordings, allowing recurring people and activities to support dependencies across distant moments.
This yields an initial pool of about 3.2k videos. 
We then use a pre-trained LLM to annotate video domains and scene composition, guiding the selection of videos across different semantic domains with recurring entities and events suitable for cross-turn dependencies.
Further collection details are provided in \autoref{app:collection}.

\noindent$\bullet$ \textbf{Dependency Planning.}
To make dependencies structural rather than incidental (Req-1), the dialogue is designed before any question is generated.
The planner samples keyframes across the video and produces a global dialogue plan with four components: 
(i) a video synopsis summarizing characters, events, and settings; 
(ii) dialogue guidelines specifying target content, capabilities to assess, and the number of turns; 
(iii) a dependency graph assigning parent turns and edge types to each dependent turn; 
and (iv) per-turn intents specifying the information to establish, the relevant video segment, and the information required from parent turns. 
The plan guides subsequent generation under two constraints: dependencies must arise from recurring video content, and the planned turns must collectively cover the full video.
More planning details are provided in \autoref{app:planning}.

\noindent$\bullet$ \textbf{Turn-wise Generation.}
Following the plan, the generator constructs the dialogue turn by turn, densely sampling frames from each turn's target segment.
For dependent turns, the segments are associated with their parents. 
It first identifies a verifiable visual fact as the reference answer, then formulates a question according to the planned intent and dependency types. Dependent questions use referring expressions without restating the information supplied by their parents. 
The supporting temporal interval is recorded as the answer's evidence window.

\noindent$\bullet$ \textbf{Turn-level Check.}
For dependent turns, the judger checks necessity by removing parent turns and rejecting questions that remain answerable from the video frames and remaining dialogue. 
It checks non-leakage by comparing each question with its parents for restated episode descriptions or disclosed answers. 
For every turn, it verifies that frames from the annotated evidence window support the reference answer and rejects questions that semantically duplicate earlier ones. 
Rejected turns are regenerated using the judger's feedback; two consecutive failures trigger replanning of the affected portion of the dialogue.
Our prompt templates for all stages are given in \autoref{app:prompt}.

\noindent$\bullet$ \textbf{Expert Review.}
Eight human experts review all dialogues that pass the automated checks against the corresponding videos. 
Turns with incorrect answers, invalid dependencies, or information leakage are revised or removed, ensuring that every retained dialogue is human-reviewed. 
Details of the dataset review and the separate human baseline evaluation are provided in \autoref{app:review}.

\subsection{Dataset Statistics}
\label{sec:stats}

\begin{figure*}[t]
\centering
\includegraphics[width=1\linewidth]{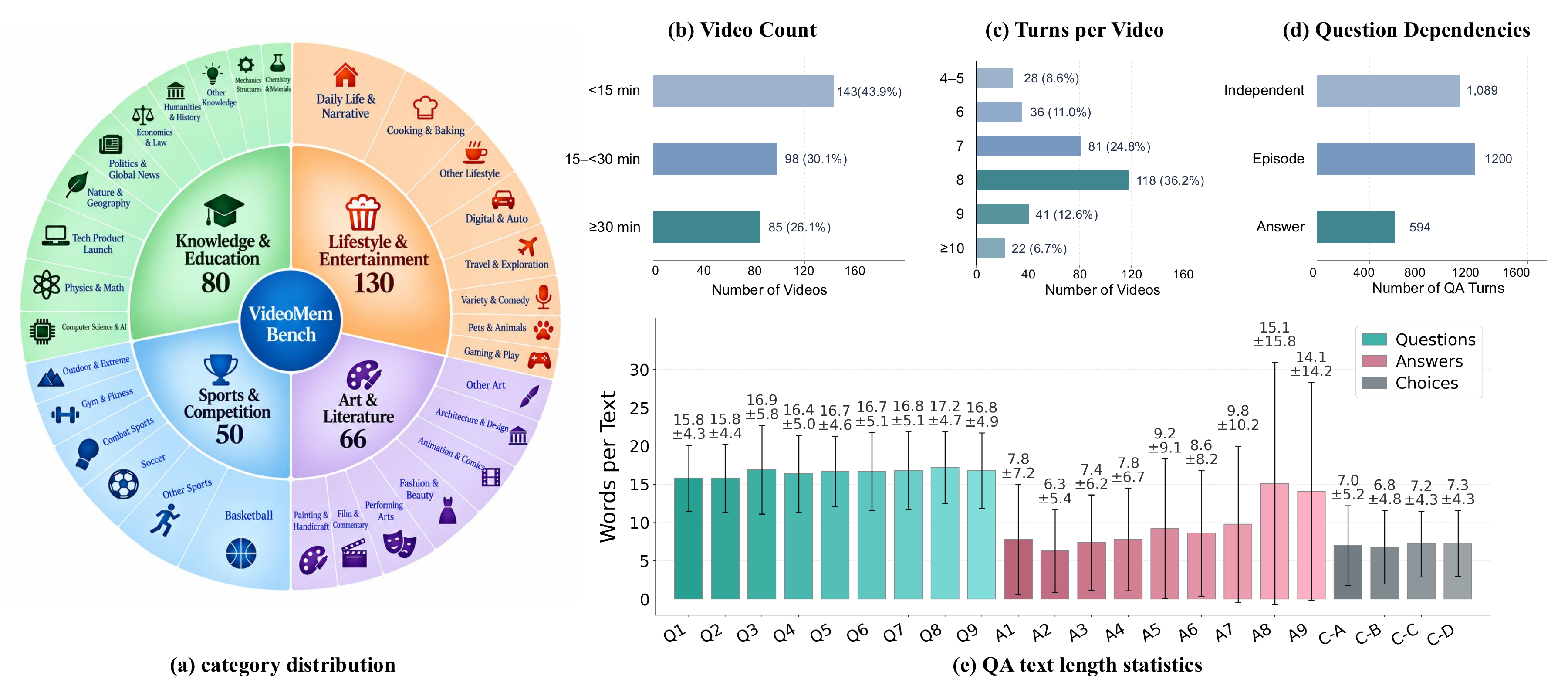}
\caption{\textbf{Dataset statistics of VideoMemBench.}
(a) Category distribution across four major domains and their fine-grained subcategories.
(b) Distribution of videos by duration.
(c) Distribution of the number of dialogue turns per video.
(d) Distribution of question dependencies, including independent questions, episodic dependencies, and answer dependencies.
(e) Text-length statistics for questions, answers, and answer choices, reported as mean $\pm$ standard deviation.}
\label{fig:dataset_statistics}
\end{figure*}

\autoref{fig:dataset_statistics} summarizes the overall statistics of VideoMemBench.
VideoMemBench contains 326 dialogues grounded in 326 videos, totaling 2,476 QA turns. 
Of these, 1,089 (43.98\%) are independent, while 1,387 (56.02\%) are dependent, with 1,794 typed dependencies in total: 1,200 Episodic and 594 Answer dependencies.
Among dependent turns, 348 have multiple parents, including 289 double-parent and 59 triple-parent turns. 
Dialogues span 4-21 turns, averaging 7.6 turns.
Among dependent turns, dependency distance ranges from one to nine turns (\eg, $T_1 \rightarrow T_5$ has distance 4), with up to three parents (\eg, $T_1 + T_4 \rightarrow T_6$ has 2 parents) and a maximum chain depth of seven (\eg, $T_1 \rightarrow T_3 \rightarrow T_4 \rightarrow T_7$ has depth 3). Further details are provided in the appendix.

\section{Experiments}
\label{sec:experiments}

\subsection{Experimental Setup}
\label{sec:setup}

\noindent \textbf{Models.}
We evaluate 25 MLLMs, including 8 closed-source models: Claude Opus 5~\citep{Claude}, Claude Sonnet 5~\citep{sonnet5}, GPT-5.6-sol~\citep{openai2026gpt56}, GPT-5.5~\citep{openai2026gpt55}, GPT-5.4~\citep{openai2026gpt54}, Gemini 3.8 Flash~\citep{google2026gemini38flash}, Gemini 3.5 Flash~\citep{google2026gemini35}, and Gemini 3.1 Pro Preview~\citep{google2026gemini31pro}, as well as 17 open-weight models: Kimi K3~\citep{Kimi}, Qwen3.5 (0.8B, 4B, 9B, 27B, 35B-A3B, 122B-A10B, 397B-A17B)~\citep{qwen35blog}, Qwen3-VL (4B, 8B, 30B-A3B, 32B, 235B-A22B)~\citep{bai2025qwen3vl}, Qwen2.5-VL (3B, 7B, 32B)~\citep{bai2025qwen25vltechnicalreport}, and Keye-VL-1.5-8B~\citep{yang2025kwai}. 

\noindent \textbf{Evaluation protocol.}
All models receive the same 128 frames sampled across the full video, with timestamps indicating their temporal positions. 
The sampled frame sequence is fixed across models.
The video frames are provided at the beginning of the conversation and remain accessible throughout the dialogue. 
Questions are presented strictly in dialogue order, with the model's own previous responses retained in the conversation history. Specifically, the input at turn $t$ is
\begin{equation}
V \,\|\, (q_1,\hat{a}_1) \,\|\, \cdots \,\|\, (q_{t-1},\hat{a}_{t-1}) \,\|\, q_t,
\end{equation}
where $V$ denotes the timestamped frame sequence, $q_i$ the question and its answer options where applicable, $\hat{a}_i$ is the model response at turn $i$, and $\|$ denotes concatenation. 

\begin{table}[t]
\centering
\caption{\textbf{Main results on VideoMemBench.} 
Models are evaluated under the self-rollout protocol.
TurnAcc: accuracy over all turns, scored individually. 
EpiDepAcc: accuracy on turns with at least one Episodic dependency. 
AnsDepAcc: accuracy on turns with at least one Answer dependency under chain crediting. 
Overall: the arithmetic mean of the three metrics.
The best and second-best results in each metric within each model group are \textbf{bold} and \underline{underlined}, respectively.}
\label{tab:main}
\small
\renewcommand{\arraystretch}{0.92}
\begin{tabular*}{\linewidth}{@{\extracolsep{\fill}}lccccc@{}}
\toprule
Model & Frames & TurnAcc (\%) & EpiDepAcc (\%) & AnsDepAcc (\%) & \multicolumn{1}{|c}{Overall (\%)} \\
\midrule
Human  & -- & 88.1 & 86.0 & 84.2 & \multicolumn{1}{|c}{86.1} \\
\midrule
\multicolumn{6}{@{}l}{\emph{Closed-source models}} \\
Claude Opus 5 & 128 & \textbf{52.3} & \textbf{49.1} & \textbf{35.7} & \multicolumn{1}{|c}{\textbf{45.7}} \\
Gemini 3.8 Flash & 128 & \underline{50.4} & 46.6 & \underline{30.3} & \multicolumn{1}{|c}{\underline{42.4}} \\
GPT-5.6-sol & 128 & 50.0 & \underline{47.8} & 27.1 & \multicolumn{1}{|c}{41.7} \\
Gemini 3.5 Flash & 128 & 49.6 & 45.5 & 29.7 & \multicolumn{1}{|c}{41.6} \\
GPT-5.5 & 128 & 48.0 & 46.5 & 27.0 & \multicolumn{1}{|c}{40.5} \\
Claude Sonnet 5 & 128 & 47.8 & 45.3 & 27.3 & \multicolumn{1}{|c}{40.2} \\
Gemini 3.1 Pro Preview & 128 & 45.7 & 44.8 & 27.0 & \multicolumn{1}{|c}{39.1} \\
GPT-5.4 & 128 & 44.4 & 42.6 & 21.9 & \multicolumn{1}{|c}{36.3} \\
\midrule
\multicolumn{6}{@{}l}{\emph{Open-weight models}} \\
Kimi K3 & 128 & \textbf{50.5} & \textbf{49.0} & \textbf{27.1} & \multicolumn{1}{|c}{\textbf{42.2}} \\
Qwen3-VL-235B-A22B & 128 & \underline{43.1} & \underline{41.6} & 18.6 & \multicolumn{1}{|c}{\underline{34.4}} \\
Qwen3.5-27B & 128 & 42.2 & 39.8 & \underline{20.9} & \multicolumn{1}{|c}{34.3} \\
Qwen3.5-397B-A17B & 128 & 42.3 & 40.1 & 20.3 & \multicolumn{1}{|c}{34.2} \\
Qwen3.5-122B-A10B & 128 & 42.4 & 39.5 & 19.9 & \multicolumn{1}{|c}{33.9} \\
Qwen3.5-35B-A3B & 128 & 38.2 & 36.3 & 18.0 & \multicolumn{1}{|c}{30.8} \\
Qwen3-VL-32B & 128 & 39.3 & 35.9 & 17.2 & \multicolumn{1}{|c}{30.8} \\
Qwen3-VL-30B-A3B & 128 & 36.4 & 35.1 & 15.8 & \multicolumn{1}{|c}{29.1} \\
Qwen3.5-9B & 128 & 34.6 & 33.4 & 12.9 & \multicolumn{1}{|c}{27.0} \\
Qwen3-VL-8B & 128 & 34.8 & 32.1 & 12.3 & \multicolumn{1}{|c}{26.4} \\
Qwen3.5-4B & 128 & 33.8 & 32.6 & 12.5 & \multicolumn{1}{|c}{26.3} \\
Qwen3-VL-4B & 128 & 32.4 & 29.4 & 12.9 & \multicolumn{1}{|c}{24.9} \\
Qwen2.5-VL-32B & 128 & 31.7 & 32.6 & 9.6 & \multicolumn{1}{|c}{24.6} \\
Qwen2.5-VL-7B & 128 & 26.2 & 24.9 & 8.4 & \multicolumn{1}{|c}{19.8} \\
Keye-VL-1.5-8B & 128 & 25.4 & 24.5 & 8.2 & \multicolumn{1}{|c}{19.4} \\
Qwen2.5-VL-3B & 128 & 23.5 & 24.4 & 5.7 & \multicolumn{1}{|c}{17.9} \\
Qwen3.5-0.8B & 128 & 21.6 & 21.3 & 5.7 & \multicolumn{1}{|c}{16.2} \\
\bottomrule
\end{tabular*}
\end{table}

\subsection{Evaluation Metrics}
\label{sec:metrics}

We use GPT-5.4 mini~\citep{openai2026gpt54mini} as the judge for open-ended responses and extract the predicted option for multiple-choice questions.
Each response is graded as correct or incorrect. 
We report three metrics with distinct evaluation subsets and scoring rules, plus an aggregate score.

\textbf{Turn Accuracy (TurnAcc).} 
TurnAcc measures accuracy over all turns. Each response is scored independently of whether the model answered its parent turns correctly.

\textbf{Episodic Dependency Accuracy (EpiDepAcc).} 
EpiDepAcc measures accuracy over turns with at least one Episodic dependency. Each response is scored independently of its parents' answer correctness, since an Episodic dependency refers to the visual context described by a parent question rather than the parent's answer.
This metric measures recall of \textit{what the model has seen}.

\textbf{Answer Dependency Accuracy (AnsDepAcc).} AnsDepAcc evaluates turns with at least one Answer dependency using \emph{chain crediting}: a turn receives credit only if its response and all responses along its Answer-dependency ancestry are correct.
Formally,
\begin{equation}
\mathrm{AnsDepAcc}
=
\frac{1}{|\mathcal{T}_{\mathrm{ans}}|}
\sum_{t \in \mathcal{T}_{\mathrm{ans}}}
\mathrm{corr}(t)
\prod_{a \in \mathrm{Anc}_{\mathrm{ans}}(t)}
\mathrm{corr}(a),
\end{equation}
where $\mathcal{T}_{\mathrm{ans}}$ contains turns with at least one Answer dependency, $\mathrm{corr}(t)\in\{0,1\}$ indicates whether the response at turn $t$ is correct, and $\mathrm{Anc}_{\mathrm{ans}}(t)$ contains all ancestors reachable by recursively following only Answer edges toward earlier turns.
Episodic edges are excluded from this traversal. 
By accounting for both current-turn errors and upstream failures across answer-dependent chains, this metric evaluates recall of \emph{what the model has said}.

Multi-dependent turns with both dependency types contribute to both EpiDepAcc and AnsDepAcc under their respective scoring rules.

\textbf{Overall.} 
Overall is the arithmetic mean of TurnAcc, EpiDepAcc, and AnsDepAcc, which is used to simply rank models. 
It aggregates complementary, overlapping evaluation subsets rather than representing accuracy on a separate set of questions.

\begin{figure*}[t]
\centering
\includegraphics[width=1\linewidth]{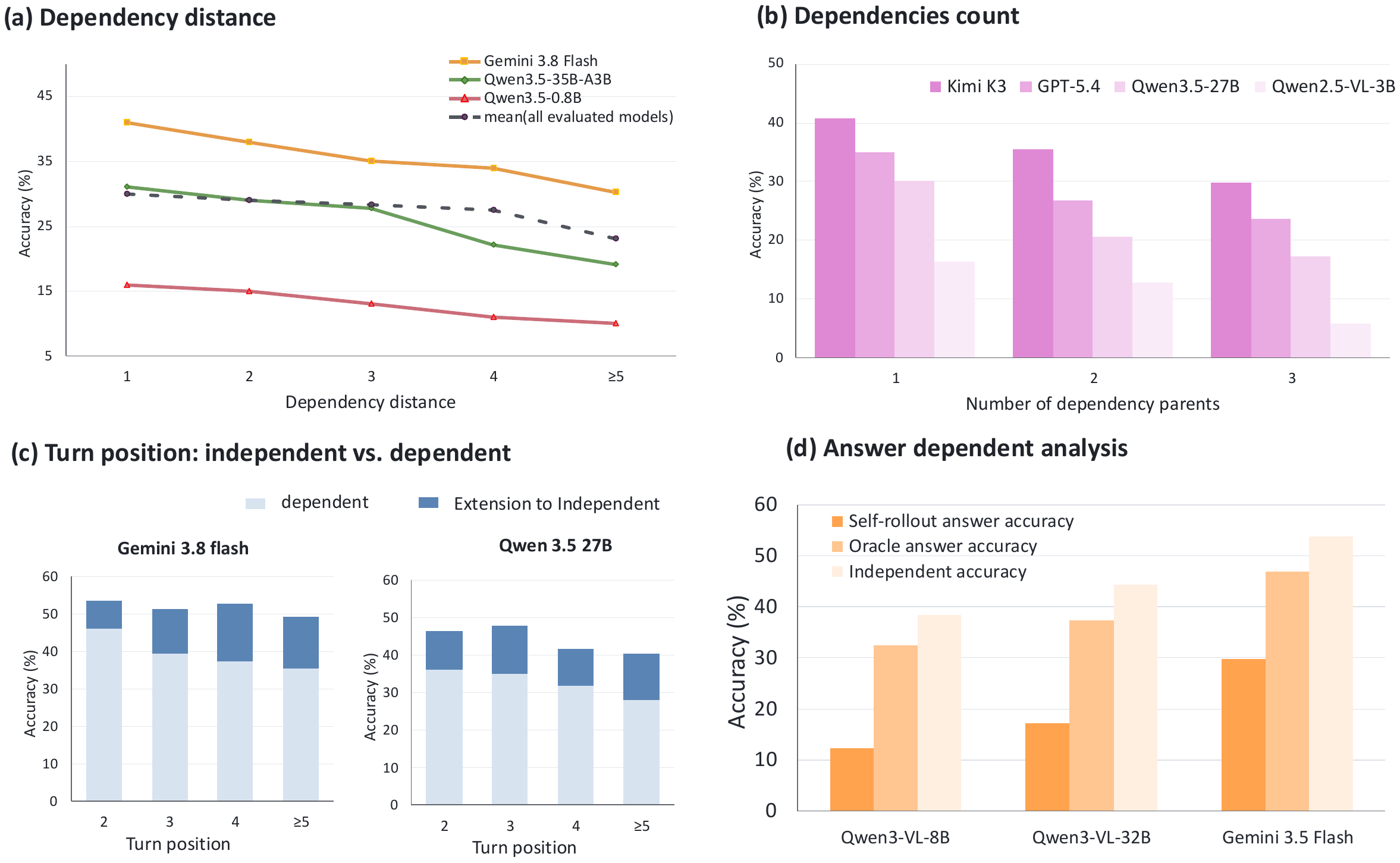}
\caption{\textbf{Analysis of factors affecting in-context video memory.} 
(a) Accuracy on dependent turns by dependency distance, measured from the current turn to its earliest parent; the dashed line shows the mean across all evaluated models.
(b) Accuracy on dependent turns by the number of parents.
(c) Independent- and dependent-turn accuracy by dialogue position for two models: the light segment shows dependent-turn accuracy, the full bar height shows independent-turn accuracy, and the dark segment represents their difference.
(d) Accuracy on answer-dependent turns under self-rollout and oracle histories. Oracle histories replace previous model responses with ground-truth answers.}
\label{fig:dependency_analysis}
\end{figure*}

\subsection{Benchmark Results}
\label{sec:main-results}

\autoref{tab:main} reports results for 25 models, including 8 closed-source and 17 open-weight MLLMs. 
Claude Opus 5 achieves the highest overall score of 45.7\%, compared with 86.1\% for human annotators. 
This gap suggests substantial room for MLLMs to improve in-context video memory.

\paragraph{Closed-source Models.}
Claude Opus 5 leads the closed-source models across all three metrics, achieving 52.3\% on TurnAcc, 49.1\% on EpiDepAcc, and 35.7\% on AnsDepAcc. Gemini 3.8 Flash and GPT-5.6-sol follow with overall scores of 42.4\% and 41.7\%, respectively, while GPT-5.4 scores lowest at 36.3\%.
Across these closed-source models, AnsDepAcc is 16.6--22.9 percentage points below TurnAcc, compared with only 3.9 percentage points for human annotators.
Since AnsDepAcc requires correctness throughout the Answer-dependency ancestry, this gap reflects both current-turn errors and accumulated upstream failures.
These results highlight the difficulty of maintaining correct answers across dependent turns in extended video conversations, even for frontier models.

\textbf{Open-weight Models.} 
Kimi K3 leads the open-weight models with an overall score of 42.2\%, ranking third overall and second on both TurnAcc and EpiDepAcc across all evaluated models.
It achieves 50.5\% on TurnAcc, 49.0\% on EpiDepAcc, and 27.1\% on AnsDepAcc. 
All other open-weight models score below GPT-5.4, the lowest-performing closed-source model (36.3\% overall).
The runner-up, Qwen3-VL-235B-A22B, scores 34.4\% overall, 7.8 percentage points below Kimi K3 and narrowly ahead of Qwen3.5-27B (34.3\%). 
Within the evaluated dense Qwen3.5 models, the overall score increases with model size, from 16.2\% at 0.8B to 34.3\% at 27B. 
At comparable scales, Qwen3-VL-32B outperforms Qwen2.5-VL-32B by 6.2 percentage points. 
These results show substantial variation among open-weight models: Kimi K3 is competitive with frontier closed-source models in our in-context video memory setting, while the other evaluated models remain behind.

\begin{table}[t]
\centering
\caption{\textbf{Independent versus dependent turns.} Indep.: accuracy over the independent turns. Dep.: mean of EpiDepAcc and AnsDepAcc. $\Delta = \mathrm{Indep.} - \mathrm{Dep.}$, measured in percentage points (pp).}
\label{tab:dep-indep}
\small
\renewcommand{\arraystretch}{0.92}
\begin{tabular*}{\linewidth}{@{\extracolsep{\fill}}lccccc@{}}
\toprule
Model & Indep. (\%) & EpiDepAcc (\%) & AnsDepAcc (\%) & Dep. (\%) & $\Delta$ (pp) \\
\midrule
Kimi K3 & 51.4 & 49.0 & 27.1 & 38.1 & 13.4 \\
GPT-5.6-sol & 50.6 & 47.8 & 27.1 & 37.5 & 13.1 \\
Claude Sonnet 5 & 50.0 & 45.3 & 27.3 & 36.3 & 13.7 \\
Qwen3.5-27B & 46.0 & 39.8 & 20.9 & 30.3 & 15.7 \\
Qwen3-VL-235B-A22B & 47.4 & 41.6 & 18.6 & 30.1 & 17.3 \\
\bottomrule
\end{tabular*}
\end{table}

\subsection{Analysis}
\label{sec:analysis}

\textbf{Dependency Distance.}
\autoref{fig:dependency_analysis}(a) groups dependent turns by the distance to their earliest parent.
Mean accuracy across all evaluated models decreases from 30.2\% at distance 1 to 23.1\% at distances of at least five turns, with a larger decline in the longest-distance group.
The model-specific trends differ: Gemini 3.8 Flash declines relatively steadily from 41.1\% to 30.2\%, whereas Qwen3.5-35B-A3B drops more sharply beyond distance 3, falling from 27.8\% to 19.2\%.
These trends suggest that longer dependencies pose greater challenges for in-context video memory, although models differ in their sensitivity to dependency distance.

\textbf{Number of Parents.}
\autoref{fig:dependency_analysis}(b) shows a consistent decline in accuracy as the number of parent turns increases across all four models.
Kimi K3 remains the strongest at every parent count, but its accuracy falls from 40.8\% with one parent to 29.8\% with three.
GPT-5.4 and Qwen3.5-27B show the same trend, falling from 34.9\% to 23.7\% and from 30.1\% to 17.3\%, respectively. 
These results suggest that integrating information from multiple earlier turns poses an additional challenge for in-context video memory, even for the strongest model in this comparison.

\textbf{Turn Position and Dependency.}
\autoref{tab:dep-indep} shows that the mean of EpiDepAcc and AnsDepAcc is 13.1--17.3 percentage points below independent-turn accuracy across the five reported models, although this gap also reflects the stricter chain-crediting rule used by AnsDepAcc.
\autoref{fig:dependency_analysis}(c) further compares independent and dependent turns at different dialogue positions.
For Gemini 3.8 Flash, dependent-turn accuracy decreases from 46.1\% at turn 2 to 35.2\% at turns 5 and beyond, while independent-turn accuracy declines more modestly, from 53.8\% to 49.2\%.
Qwen3.5-27B exhibits a similar pattern: dependent-turn accuracy falls from 36.2\% to 28.1\%, compared with a decline from 46.1\% to 40.3\% on independent turns.
Although independent-turn accuracy fluctuates at intermediate positions, dependent-turn accuracy declines consistently for both models.
These patterns suggest that later dialogue positions are associated with greater difficulty, particularly when answering requires information from earlier turns.

\textbf{Effect of Answer History.}
\autoref{fig:dependency_analysis}(d) compares accuracy on answer-dependent turns under self-rollout and oracle histories, with independent-turn accuracy as a reference. 
Oracle histories replace previous model responses with reference answers.
All three models benefit substantially: Qwen3-VL-8B improves from 12.3\% to 32.4\%, Qwen3-VL-32B from 17.2\% to 37.3\%, and Gemini 3.5 Flash from 29.7\% to 47.2\%.
These gains indicate that errors in earlier responses are an important source of difficulty for subsequent answer-dependent questions.
Nevertheless, oracle accuracy remains below independent-turn accuracy, suggesting that correct history alone is insufficient for reliable in-context video memory.

We further provide frame-count ablations with 16, 32, and 64 sampled frames in Appendix~\ref{app:frame-count} and additional dialogue examples in Appendix~\ref{app:examples}.

\section{Conclusion}

We introduce VideoMemBench to evaluate video memory through explicit Episodic and Answer dependencies, distinguishing what a model has seen from what it has said, with automated verifications and human reviews. 
Evaluation of over 20 MLLMs shows that the best model achieves only 45.7\% overall, with lower accuracy on longer-range and multi-parent dependencies.
These findings highlight the difficulty of maintaining and using dialogue context in video conversations.
We hope VideoMemBench supports the development of video MLLMs that reliably recall and build on earlier observations and responses.

\clearpage

\bibliography{iclr2027_conference}
\bibliographystyle{iclr2027_conference}

\clearpage

\appendix

\section{Video Collection Details}
\label{app:collection}

\paragraph{Video sources.}
We assemble an initial pool of approximately 3.2k videos from five existing video-understanding benchmarks and the web. For web collection, we search YouTube using queries spanning sports (``full match'' and ``match replay''), talks and lectures (``keynote'' and ``course lecture''), tutorials (``full lesson''), film recaps (``movie recap'' and ``film summary''), and cooking (``cooking lesson'' and ``chef pov'').

\paragraph{Domain taxonomy.}
The videos are organized into four broad categories, each comprising finer-grained domains:

\begin{itemize}
    \item \textbf{Lifestyle \& Entertainment (130 videos).}
    This category covers daily activities and entertainment, including Daily Life \& Narrative, Cooking \& Baking, Digital \& Auto, Travel \& Exploration, Variety \& Comedy, Pets \& Animals, Gaming \& Play, and Other Lifestyle.

    \item \textbf{Knowledge \& Education (80 videos).}
    This category spans scientific, technical, and social topics: Computer Science \& AI, Physics \& Math, Tech Product Launch, Nature \& Geography, Politics \& Global News, Economics \& Law, Humanities \& History, Mechanics Structures, and Chemistry \& Materials, together with Other Knowledge.

    \item \textbf{Art \& Literature (66 videos).}
    This category includes Painting \& Handicraft, Film \& Commentary, Performing Arts, Fashion \& Beauty, Animation \& Comics, Architecture \& Design, and Other Art.

    \item \textbf{Sports \& Competition (50 videos).}
    This category covers Basketball, Soccer, Combat Sports, Gym \& Fitness, Outdoor \& Extreme, and Other Sports.
\end{itemize}

\section{Dependency Planning Details}
\label{app:planning}

Before question generation, the planner examines keyframes sampled across the video and produces a complete dialogue specification. The plan contains four components:

\paragraph{Video synopsis.}
The synopsis summarizes the main characters, events, and settings, providing the generator with global context for interpreting individual video segments.

\paragraph{Dialogue guidelines.}
The planner identifies which parts of the video to cover, which capabilities to assess, and how many turns to construct. Target capabilities include entity recall, event continuation, and cross-scene character tracking.

\paragraph{Dependency graph.}
The planner constructs a directed acyclic graph whose nodes represent dialogue turns. Each dependent turn is assigned one or more earlier parent turns, with each incoming edge labeled as Episodic or Answer. Specifying the graph before generation allows dependencies to be coordinated across the dialogue.

\paragraph{Per-turn intents.}
For each turn, the planner specifies what information it should establish and which video segment supports it. For dependent turns, the intent additionally identifies the information required from each parent, giving the generator an explicit basis for constructing the question.

The plan must satisfy two constraints. First, dependencies must be motivated by recurring entities, events, or details in the video, rather than imposed between unrelated topics. Second, the selected segments must span the video rather than concentrate on a local clip. The complete plan is passed to the generator as the specification for turn-wise generation.

\section{Prompt Templates}
\label{app:prompt}

We report the prompt templates used to construct and evaluate VideoMemBench, following the pipeline in \autoref{fig:construction_pipeline}.
During initial video curation, a pre-trained LLM first assigns each collected video a semantic domain and scene tags.
In Stage~1, a Planner LLM watches each curated video and produces a dialogue-construction plan: a video introduction, a dialogue suggestion, the number of turns, a per-turn intent with its evidence window, and a typed dependency graph.
In Stages~2 and~3, a Generator LLM writes the dialogue turn by turn according to the plan, and a Judge LLM applies three turn-level checks (dependency necessity, non-leakage, and visual grounding); turns that fail are returned to the Generator LLM with the judge's feedback for regeneration.
Each completed dialogue is then checked against the planner's dependency graph (plan fidelity) and reviewed by human experts.
We preserve runtime placeholders such as \texttt{\{video\_frames\}}, \texttt{\{plan\}}, and \texttt{\{parent\_turns\}}.
In all prompts, an \emph{answer} dependency means the child turn needs the parent turn's correct answer, and an \emph{episodic} dependency means the child turn only needs to relocate the same scene or entity as the parent turn.

\paragraph{Video curation.}
\leavevmode\par
\begin{promptbox}{Labeler LLM (Gemini 3 Flash): Domain and Scene Tagging}
You are tagging videos for a multi-turn video memory benchmark.
You will see frames sampled uniformly across one video, each preceded by a <MM:SS> time tag.

Assign exactly ONE semantic domain from this list:
{domain_taxonomy}

Then describe the scene structure of the video.

(*@\textbf{Rules}@*):
- Choose the domain from what is shown on screen, not from the video title or file name.
- A scene is a continuous stretch with the same location and the same main participants.
  List every distinct scene with its time window.
- Mark a video as unsuitable if it has fewer than 3 visually distinct scenes, if most frames are
  static text or slides, or if the content cannot be verified from the frames alone.
- Do not guess what happens between sampled frames.

Output ONLY JSON:
{
  "domain": string,                    // one entry from the domain list
  "domain_confidence": number,         // 0..1
  "scenes": [
    {"scene_id": "S1", "t_start": number, "t_end": number,
     "location": string, "main_entities": [string, ...], "summary": string}
  ],
  "num_distinct_scenes": int,
  "suitable": true|false,
  "unsuitable_reason": string|null
}
\end{promptbox}

\paragraph{Stage 1: planning.}
\leavevmode\par
\begin{promptbox}{Planner LLM (GPT 5.5): Dialogue-Construction Plan}
You are planning a multi-turn question-answering dialogue about one video for a benchmark
that tests in-context video memory: later questions should only be answerable by remembering
what earlier turns established.
You will see frames sampled across the whole video, each preceded by a <MM:SS> time tag, and
the video's domain and scene list from the labeler.

Do NOT write questions or answers. Produce a PLAN that another model will follow.

(*@\textbf{Step 1: Video introduction}@*)
Write 2-4 sentences describing what happens in the video, in order, naming the main entities
and scene changes. Only describe what is visible in the frames.

(*@\textbf{Step 2: Dialogue suggestion}@*)
Decide how many turns the dialogue should have (between {min_turns} and {max_turns}) and how
it should move through the video. Pick at least three scene clusters that are far apart in time
and describe the route: what is established first, what is revisited next, what is tracked last.
List the memory abilities the dialogue will exercise, chosen from: entity recall, event
continuation, clue-based reasoning, cross-scene tracking, spatial localization.

(*@\textbf{Step 3: Intent for every turn}@*)
For each turn, give:
- turn_id (T1, T2, ...)
- question_type: multi_choice_question, true_or_false_question, or open_structured
- dependency: independent, or a list of parent turns, each typed as "answer" or "episodic"
- evidence window [t_start, t_end] in seconds where the answer is visible
- intent: one sentence saying what the turn should establish or test

(*@\textbf{Step 4: Dependency structure}@*)
Design the dependency graph with these rules:
- "answer": the child cannot be answered without the parent's CORRECT answer.
  Test: if the parent were answered wrongly, would the child's correct answer change? Only then use "answer".
  The parent of an "answer" edge may use any supported question type, provided that its answer supplies information required by the child.
- "episodic": the child only needs to return to the same scene, moment, or entity as the parent.
  The parent's answer being wrong does not change the child's answer.
- Parents must come earlier than the child. The graph must have no cycles.
- More than half of the turns must have at least one parent.
- At least one turn must have two or three parents.
- At least one edge must span two or more turns (not only adjacent turns).
- T1 must be independent and may use any supported question type.
- Evidence windows must cover the beginning, middle, and end of the video.
- Every edge must be genuine: the child's target must be hard to describe without the parent.
  Do not connect turns that are about unrelated things.

Input:
domain: {domain}
scenes: {scenes}
frames: {video_frames}

Output ONLY JSON:
{
  "video_intro": string,
  "dialogue_suggestion": {
    "num_turns": int,
    "scene_route": string,
    "abilities": [string, ...]
  },
  "turns": [
    {"turn_id": "T1", "question_type": string,
     "dependency": "independent" | [{"parent": "T#", "type": "answer"|"episodic"}, ...],
     "evidence": {"t_start": number, "t_end": number},
     "intent": string}
  ],
  "dependency_graph": [{"from": "T#", "to": "T#", "type": "answer"|"episodic"}, ...]
}
\end{promptbox}

\paragraph{Stage 2: generation.}
\leavevmode\par
\begin{promptbox}{Generate LLM (Gemini 3.5 Flash): Turn Generation}
You are writing one turn of a multi-turn video QA dialogue, following a plan.
You will see frames from the video with <MM:SS> time tags, the plan, and the turns already written.

Write turn {turn_id} so that it matches its planned intent, question type, dependency, and
evidence window exactly.

(*@\textbf{Build the answer first}@*)
1. Look at the frames inside the planned evidence window and pick ONE specific fact you can
   clearly see (an identity, position, count, color, action, text, or change).
2. That fact is the answer. If you cannot see a clear fact that serves the intent, say so in
   "plan_issue" instead of guessing.
3. Write a short question (one sentence, at most two) that asks for that fact.
4. For multiple choice, write 4 options (A-D) of the same kind, each under 15 words.
   Every wrong option should be plausible for this video and wrong for exactly one reason.
5. Write answer_reasoning: 1-2 sentences naming what in the frames makes the answer correct.

(*@\textbf{Dependencies}@*)
- Refer to each parent turn inside the sentence by its turn id, for example
  "the player identified in T1" or "the shelf asked about in T3".
- For an "answer" parent, the question must be impossible to answer without knowing which
  option was correct in the parent. Refer to that result abstractly; never state it.
- For an "episodic" parent, the question returns to the same scene or entity. Give just enough
  visual anchors (appearance, location, nearby landmark) to find it again, but do not describe
  the parent's scene so fully that the parent becomes unnecessary.
- Never repeat a parent's correct option, and never use words that make it obvious.
  The answering model will see earlier questions but not earlier correct answers.
- Do not add dependencies that are not in the plan, and do not drop planned ones.

(*@\textbf{Wording}@*)
- No exact timestamps in the question; use scene descriptions ("when the scoreboard first appears").
- One question per turn. Natural language, no placeholders, no labels like "object A".
- true_or_false_question: options are exactly A = "Yes", B = "No".
- open_structured: no options; give a short reference answer and 1-3 acceptable paraphrases.

Input:
plan: {plan}
turns written so far: {previous_turns}
turn to write: {turn_plan}
frames: {video_frames}

Output ONLY JSON:
{
  "turn_id": "T#",
  "type": "multi_choice_question"|"true_or_false_question"|"open_structured",
  "dependency": {"depends_on": ["T#", ...], "dep_type": {"T#": "answer"|"episodic"}} | null,
  "question": string,
  "options": {"A": string, "B": string, "C": string, "D": string} | null,
  "answer": string,
  "acceptable_answers": [string, ...] | null,
  "answer_reasoning": string,
  "evidence": [{"t_start": number, "t_end": number}],
  "plan_issue": string|null
}
\end{promptbox}

\begin{promptbox}{Generator LLM (Gemini 3.5 Flash): Regeneration of a Failed Turn}
A turn you wrote was rejected by a quality check. Rewrite it so it passes, keeping everything
the plan requires.

You will see the plan entry for this turn, the turns written so far, the rejected turn, the
name of the failed check, and the judge's reason.

(*@\textbf{How to fix each failed check}@*)
- dependency_necessity: the turn could be answered without its parent(s). Remove the details
  that let the target be found on its own, and refer to the parent turn instead, so that the
  parent is genuinely required.
- non_leakage: the question re-described the parent's episode or revealed the parent's answer.
  Replace those words with a neutral reference to the parent turn by its turn id.
- visual_grounding: the frames in the evidence window do not support the answer. Pick a fact
  that IS visible in that window and rebuild the question and answer around it.

(*@\textbf{Keep fixed}@*): turn_id, question type, planned dependency and its types, evidence window.
If the turn cannot be fixed without changing the plan, return the turn unchanged and explain
in "plan_issue".

Input:
plan entry: {turn_plan}
turns written so far: {previous_turns}
rejected turn: {turn}
failed check: {check_name}
judge reason: {judge_reason}
frames: {video_frames}

Output ONLY JSON with the same schema as Turn Generation.
\end{promptbox}

\paragraph{Stage 3: turn-level checks.}
\leavevmode\par
\begin{promptbox}{Judge LLM (Claude Opus 4.8): Dependency Necessity Probe}
You are answering one question about a video. Some earlier turns of the conversation have been
removed on purpose. Answer only if the frames and the remaining conversation are enough.

You will see frames from the video with <MM:SS> time tags, the conversation so far with some
turns removed, and one question.

(*@\textbf{Rules}@*):
- If the question refers to an earlier turn that is not shown, you do not know what that turn
  established. Do not guess what it said.
- If you can still identify the answer with confidence from the frames alone, give it.
- If the answer depends on information you do not have, output "ABSTAIN".

Input:
frames: {video_frames}
conversation (parent turns removed): {history_without_parents}
question: {question}
options: {options}

Output ONLY JSON:
{"answer": string, "confidence": number, "why": string}
// answer: an option letter, a short answer for open questions, or "ABSTAIN"

Decision rule (applied by the pipeline, not shown to the probe): the turn is REJECTED if the probe
answers correctly with confidence >= 0.5, because then the memory chain was decorative.
\end{promptbox}

\begin{promptbox}{Judge LLM (Claude Opus 4.8): Non-Leakage Check}
You are checking whether a later question gives away its parent turns.
You will see the later question next to each of its parent turns (question, options, and correct
answer). You do not see the video.

A question LEAKS if either condition holds:
1. Answer leak: it states, paraphrases, or reveals any part of a
parent's correct answer, even if the parent is still needed to
recover the remaining information.
2. Episode re-description: it repeats identifying details of the
scene, moment, or entity established by an episodic parent,
rather than referring to it without restating those details.

Generic references such as "the player identified in T1" or
"the moment discussed in T3" are allowed.

For example, if a parent answer is "the flour in the red bowl",
"How much of the flour identified earlier is used?" leaks the
ingredient, even if the parent is still needed to identify the bowl.
Leakage does not require making the parent unnecessary.

Input:
later question: {question}
dependency types: {dep_type}
parent turns: {parent_turns}

Output ONLY JSON:
{
  "leak": true|false,
  "answer_leak_parents": ["T#", ...],
  "redescribed_parents": ["T#", ...],
  "offending_text": [string, ...],
  "reason": string
}
\end{promptbox}

\begin{promptbox}{Judge LLM (Claude Opus 4.8): Visual Grounding Check}
You are checking whether a question and its reference answer are supported by the video.
You will see ONLY the frames from the turn's annotated evidence window, each with a <MM:SS> tag,
and the question-answer pair.

(*@\textbf{Steps}@*):
1. Before reading the reference answer, answer the question yourself from these frames.
2. Compare your answer with the reference answer.
3. Check that the question itself contains no facts that contradict the frames (wrong numbers,
   text, colors, or objects).
4. For multiple choice, check that exactly one option is correct in these frames.

The turn is SUPPORTED only if the frames clearly show the fact the reference answer states and
no other option is also correct. If the key object or action is not visible in these frames,
it is NOT supported, even if you believe it happens elsewhere in the video.

Input:
frames (evidence window): {evidence_frames}
question: {question}
options: {options}
reference answer: {answer}

Output ONLY JSON:
{
  "supported": true|false,
  "my_answer": string,
  "support_ts": [string, ...],     // <MM:SS> tags of frames showing the fact
  "question_factual_errors": [string, ...],
  "multiple_correct_options": true|false,
  "reason": string
}
\end{promptbox}

\paragraph{Dialogue-level check.}
\leavevmode\par
\begin{promptbox}{Judge LLM (Claude Opus 4.8): Plan Fidelity Check}
You are checking whether a finished dialogue was built the way it was planned.
You will see the planner's per-turn intents and dependency graph, and the completed dialogue.
You do not see the video.

For every turn, read the question and decide which earlier turns it actually needs, and for each
one whether it needs the parent's correct ANSWER or only the same scene or entity (EPISODIC).
Base this on the question text, not on the dependency fields the generator wrote.

Then compare this realized graph with the planned graph:
- missing edge: planned but the question does not actually need that parent
- extra edge: the question needs a parent that the plan did not include
- retyped edge: planned as "answer" but only episodic is needed, or the reverse
Also flag any turn whose question does not serve its planned intent.

The dialogue is REJECTED if any edge is missing, extra, or retyped, or if any turn does not
serve its intent.

Input:
planned turns: {plan_turns}
planned dependency graph: {dependency_graph}
completed dialogue: {dialogue}

\end{promptbox}

\begin{figure*}[t]
\centering
\includegraphics[width=1\linewidth]{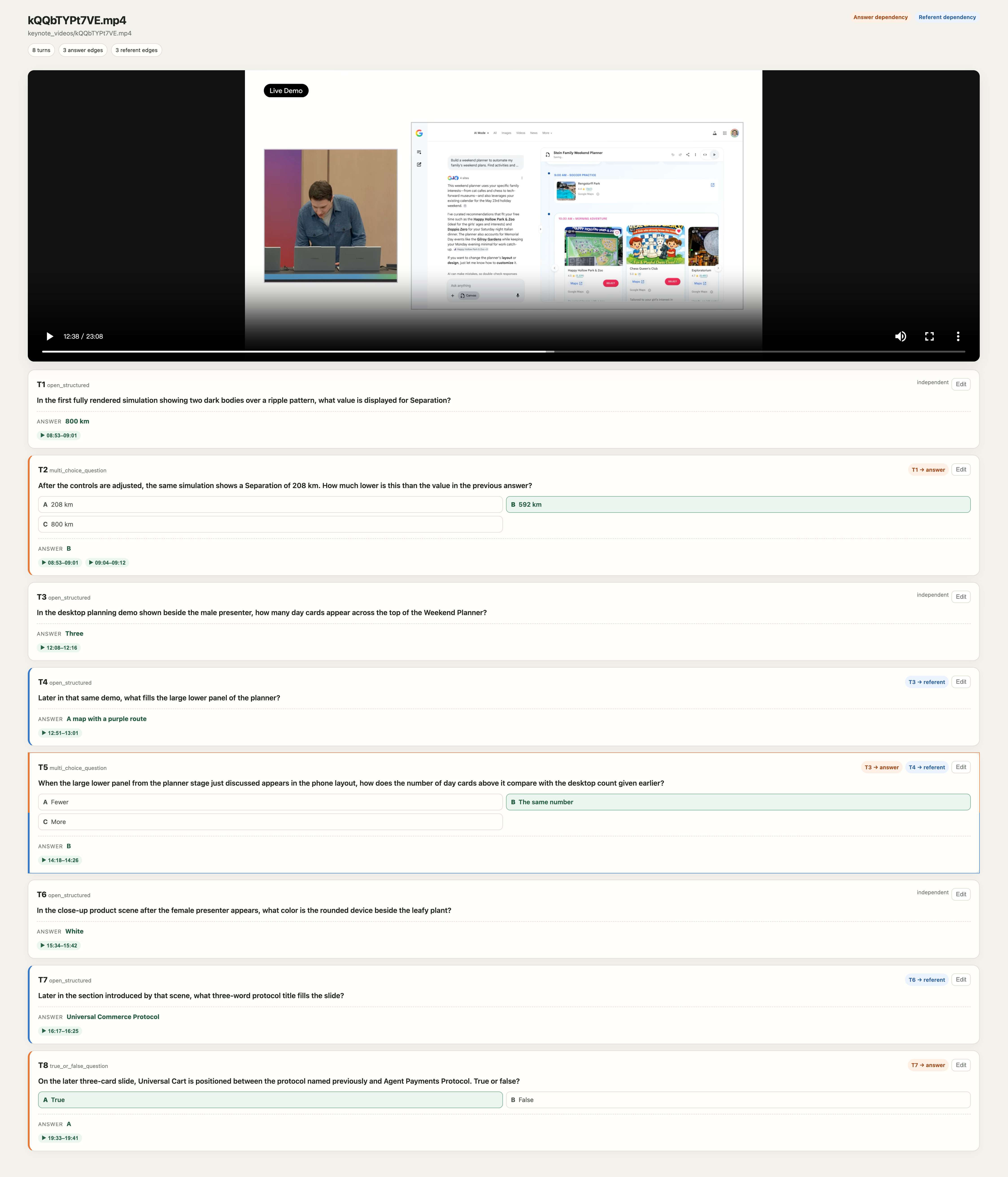}
\caption{\textbf{The expert review interface.} 
For each dialogue, the expert replays the video in the top panel, where the evidence window of the current turn is marked on the timeline, and reads the dialogue turn by turn, where the parent turns and the type of each dependency are highlighted. 
The expert answers the current question before the reference answer is revealed, then marks the turn as \emph{correct}, \emph{answer wrong}, \emph{dependency not required}, \emph{wording leaks a parent}, or \emph{duplicate}, and edits the question, the options, or the answer in place. 
A dialogue is retained only after every turn has been marked correct or revised, and the interface records each decision together with the time spent on the turn.
Referent here means episodic dependency. Please zoom in to see the details.}
\label{fig:annotation_interface}
\end{figure*}

\begin{table}[t]
\centering
\caption{\textbf{Frame-count ablation on VideoMemBench under self-rollout.}
Each model is evaluated with 16, 32, and 64 uniformly sampled frames per video, using the same metrics as the 128-frame evaluation in \autoref{tab:main}.
Models are ranked by the overall score within each frame budget, with the best result for each metric shown in \textbf{bold}.}
\label{tab:frame-ablation}
\footnotesize
\renewcommand{\arraystretch}{0.92}
\begin{tabular*}{\linewidth}{@{\extracolsep{\fill}}lcccc|c@{}}
\toprule
Model & Frames & TurnAcc (\%) & EpiDepAcc (\%) & AnsDepAcc (\%) & Overall (\%) \\
\midrule
Qwen3-VL-235B-A22B & 16 & 33.7 & 32.3 & 11.7 & \textbf{25.9} \\
Qwen3.5-122B-A10B & 16 & 33.2 & 32.4 & \textbf{11.9} & 25.8 \\
Qwen3.5-27B & 16 & \textbf{33.9} & \textbf{33.7} & 8.2 & 25.3 \\
Qwen3-VL-32B & 16 & 30.6 & 28.9 & 10.5 & 23.3 \\
Qwen3-VL-30B-A3B & 16 & 29.6 & 30.6 & 7.6 & 22.6 \\
Qwen3.5-35B-A3B & 16 & 30.7 & 28.1 & 9.0 & 22.6 \\
Qwen3-VL-8B & 16 & 27.5 & 27.7 & 6.3 & 20.5 \\
Qwen3.5-9B & 16 & 28.0 & 26.5 & 6.8 & 20.5 \\
Qwen3-VL-4B & 16 & 26.7 & 26.5 & 8.0 & 20.4 \\
Keye-VL-1.5-8B & 16 & 22.0 & 22.5 & 4.7 & 16.4 \\
Qwen2.5-VL-7B & 16 & 21.4 & 21.3 & 3.7 & 15.5 \\
Qwen2.5-VL-3B & 16 & 20.2 & 20.1 & 3.7 & 14.7 \\
Qwen3.5-0.8B & 16 & 17.8 & 18.0 & 3.9 & 13.2 \\
\midrule
Qwen3-VL-235B-A22B & 32 & \textbf{36.3} & \textbf{37.8} & \textbf{14.5} & \textbf{29.5} \\
Qwen3.5-122B-A10B & 32 & 35.0 & 35.2 & 10.7 & 27.0 \\
Qwen3.5-35B-A3B & 32 & 34.5 & 34.1 & 12.3 & 27.0 \\
Qwen3.5-27B & 32 & 34.9 & 34.2 & 11.7 & 26.9 \\
Qwen3-VL-32B & 32 & 34.1 & 32.6 & 13.1 & 26.6 \\
Qwen3-VL-30B-A3B & 32 & 32.1 & 32.5 & 8.0 & 24.2 \\
Qwen3.5-9B & 32 & 29.8 & 30.4 & 8.6 & 22.9 \\
Qwen3-VL-8B & 32 & 29.4 & 29.4 & 9.2 & 22.7 \\
Qwen3-VL-4B & 32 & 29.0 & 29.0 & 8.8 & 22.3 \\
Qwen2.5-VL-7B & 32 & 23.2 & 23.7 & 4.9 & 17.3 \\
Keye-VL-1.5-8B & 32 & 23.3 & 23.1 & 5.1 & 17.2 \\
Qwen2.5-VL-3B & 32 & 22.6 & 23.4 & 4.7 & 16.9 \\
Qwen3.5-0.8B & 32 & 20.1 & 21.8 & 4.7 & 15.5 \\
\midrule
Qwen3-VL-235B-A22B & 64 & \textbf{40.3} & \textbf{39.3} & \textbf{15.6} & \textbf{31.7} \\
Qwen3.5-122B-A10B & 64 & 37.8 & 36.5 & 15.2 & 29.8 \\
Qwen3.5-27B & 64 & 37.9 & 36.7 & 14.8 & 29.8 \\
Qwen3-VL-32B & 64 & 37.8 & 36.1 & 14.5 & 29.5 \\
Qwen3.5-35B-A3B & 64 & 37.0 & 36.7 & 13.5 & 29.1 \\
Qwen3-VL-30B-A3B & 64 & 33.6 & 32.9 & 11.5 & 26.0 \\
Qwen3-VL-8B & 64 & 32.7 & 32.0 & 11.9 & 25.5 \\
Qwen3.5-9B & 64 & 33.4 & 33.2 & 9.6 & 25.4 \\
Qwen3-VL-4B & 64 & 31.3 & 30.3 & 12.1 & 24.5 \\
Keye-VL-1.5-8B & 64 & 24.5 & 24.9 & 6.3 & 18.5 \\
Qwen2.5-VL-7B & 64 & 24.1 & 22.7 & 6.4 & 17.7 \\
Qwen2.5-VL-3B & 64 & 22.5 & 22.5 & 5.1 & 16.7 \\
Qwen3.5-0.8B & 64 & 21.0 & 21.1 & 4.7 & 15.6 \\
\bottomrule
\end{tabular*}
\end{table}

\section{Human Review Details}
\label{app:review}

\subsection{Dataset Review}

We conduct a human review of all constructed multi-turn dialogues and dependency annotations before finalizing VideoMemBench. Eight experts participate in this review. As shown in \autoref{fig:annotation_interface}, our web interface presents the video alongside the dialogue in turn order. Each turn displays its question, reference answer, supporting timestamps, question type, and dependency links to earlier turns.

Reviewers verify five aspects:
(1) \textbf{answer correctness}, whether the reference answer is supported by the video;
(2) \textbf{temporal grounding}, whether the annotated evidence intervals contain the required visual evidence;
(3) \textbf{dependency validity}, whether the annotated parent turns and dependency types match the question's intended references;
(4) \textbf{necessity}, whether the question becomes unanswerable without its parent turns, even with access to the video and the remaining dialogue; and
(5) \textbf{non-leakage}, whether the question restates information that should instead be recalled from its parents.
For turns with multiple parents, reviewers additionally assess whether each parent contributes indispensable information.

\subsection{Human Baseline Evaluation}

The human baseline in \autoref{tab:main} is evaluated separately from dataset review, using the same web interface configured for evaluation.
In this mode, annotators see only the inputs available to models, with reference answers, evidence annotations, and dependency labels hidden. 
Twenty annotators evaluate all 326 dialogues, with each dialogue assigned to two annotators who answer independently. 
This yields 652 complete dialogue evaluations: 12 annotators each evaluate 33 dialogues, and the remaining eight each evaluate 32. 
Assignments are balanced across video categories, dialogue lengths, and dependency complexity. 
Where possible, annotators are assigned dialogues they did not previously review.

\textbf{Evaluation inputs.}
Annotators receive the same 128 timestamped frames as the models, without access to the full video, reference answers, annotated evidence intervals, or dependency labels. Questions and answer options, where applicable, are presented sequentially. Each annotator can access the preceding questions and their own previous responses, but not the other annotator's answers. Dialogues are evaluated in full rather than divided among annotators, preserving a separate response history for each evaluation.

\textbf{Scoring and aggregation.}
Human responses are graded using the same correctness criteria as model responses. TurnAcc, EpiDepAcc, and AnsDepAcc follow the definitions in \autoref{sec:metrics}. For AnsDepAcc, chain credit is computed separately within each annotator's dialogue: a turn receives credit only if the current response and all of its Answer-dependency ancestors are correct. Scores are then aggregated across both evaluations of every dialogue, with equal weight given to each evaluation. Responses from different annotators are not combined through majority voting. Overall is the arithmetic mean of the three aggregated metrics.

\section{Additional Results and Examples}

\subsection{Effect of Frame Count}
\label{app:frame-count}
We evaluate 13 open-weight models with 16, 32, and 64 uniformly sampled frames per video under the same self-rollout protocol and scoring rules as the main experiments. \autoref{tab:frame-ablation} reports the results, complementing the 128-frame evaluation in \autoref{tab:main}.

\textbf{Performance across frame budgets.}
Increasing the number of frames generally improves performance. All 13 models achieve higher Overall scores with 64 frames than with 16 frames. Qwen3-VL-235B-A22B ranks first at all three frame budgets, improving from 25.9\% to 29.5\% and 31.7\% with 16, 32, and 64 frames, respectively. Its score rises further to 34.4\% with 128 frames. Qwen3.5-27B similarly improves from 25.3\% at 16 frames to 29.8\% at 64 frames and 34.3\% at 128 frames. These results suggest that denser temporal sampling provides useful visual evidence for answering questions throughout the dialogue.

\textbf{Variation across models.}
The gains from additional frames vary considerably. From 16 to 64 frames, Qwen3-VL-32B improves by 6.2 percentage points overall, compared with 2.1 points for Keye-VL-1.5-8B. Increasing the frame budget from 32 to 64 improves Overall for 12 of the 13 models, with Qwen2.5-VL-3B showing a slight decrease from 16.9\% to 16.7\%. Qwen3.5-0.8B gains only 0.1 percentage points over the same interval. Thus, additional visual input generally benefits performance, but does not yield uniform gains across models.

\textbf{Answer-dependent performance.}
AnsDepAcc remains below both TurnAcc and EpiDepAcc for every model at every frame budget in this ablation. For Qwen3-VL-235B-A22B, increasing the frame count from 16 to 64 raises AnsDepAcc from 11.7\% to 15.6\%, while Qwen3.5-27B improves from 8.2\% to 14.8\%. With 128 frames, their AnsDepAcc scores increase further to 18.6\% and 20.9\%, respectively. These gains show that additional visual evidence can improve correctness across answer-dependent chains, although the metric remains sensitive to both current-turn errors and upstream failures.

\subsection{Additional Dataset Examples}
\label{app:examples}
\begin{figure*}[t]
\centering
\includegraphics[width=\linewidth,height=0.90\textheight,keepaspectratio]{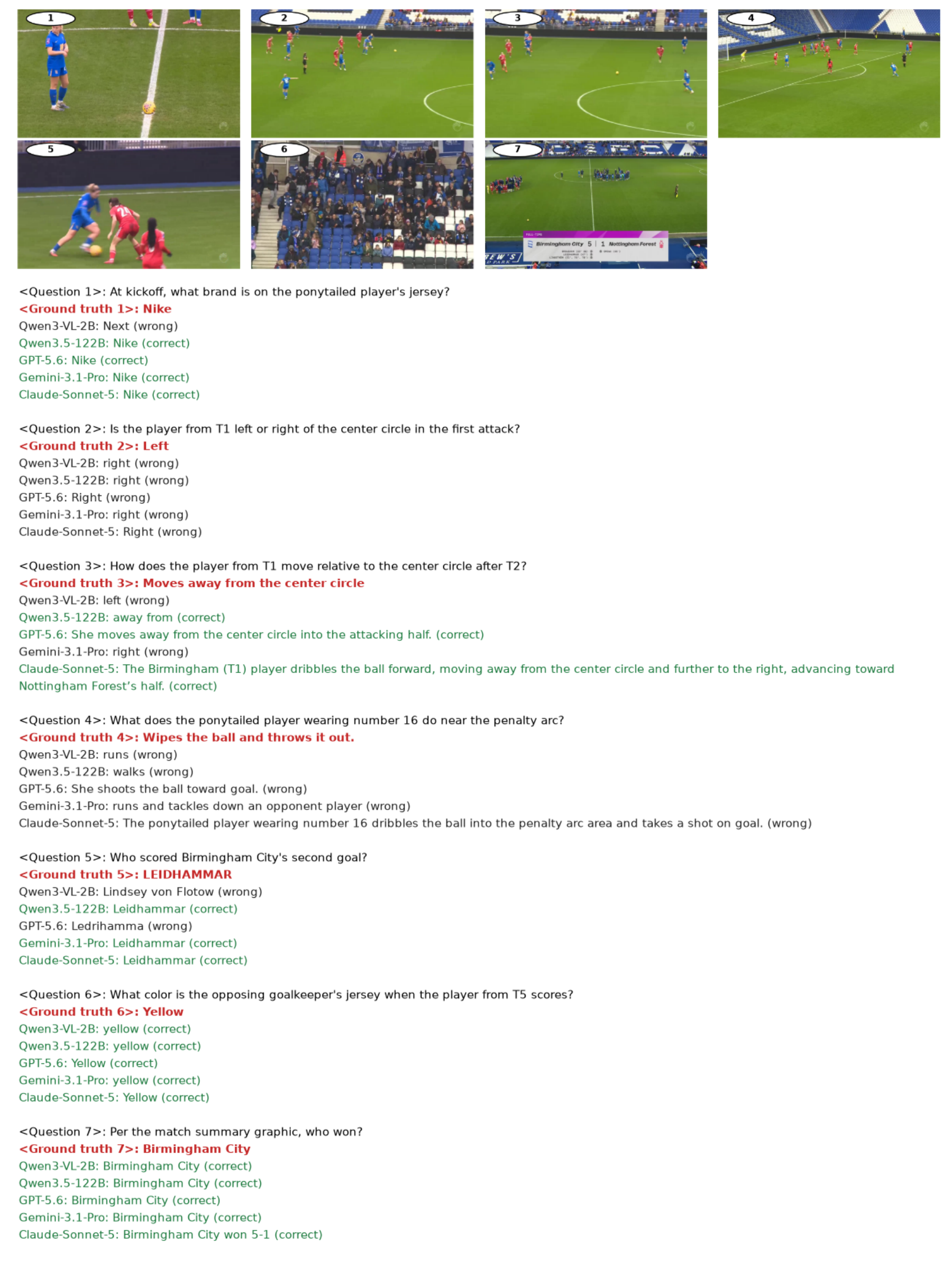}
\caption{\textbf{A soccer dialogue example from VideoMemBench.}
Seven dialogue turns cover player identification, spatial tracking, and match events, with later questions referring to previously discussed players. Reference answers are shown in red, and model responses marked correct at the turn level are shown in green.}
\label{fig:example_soccer}
\end{figure*}

\begin{figure*}[t]
\centering
\includegraphics[width=\linewidth,height=0.90\textheight,keepaspectratio]{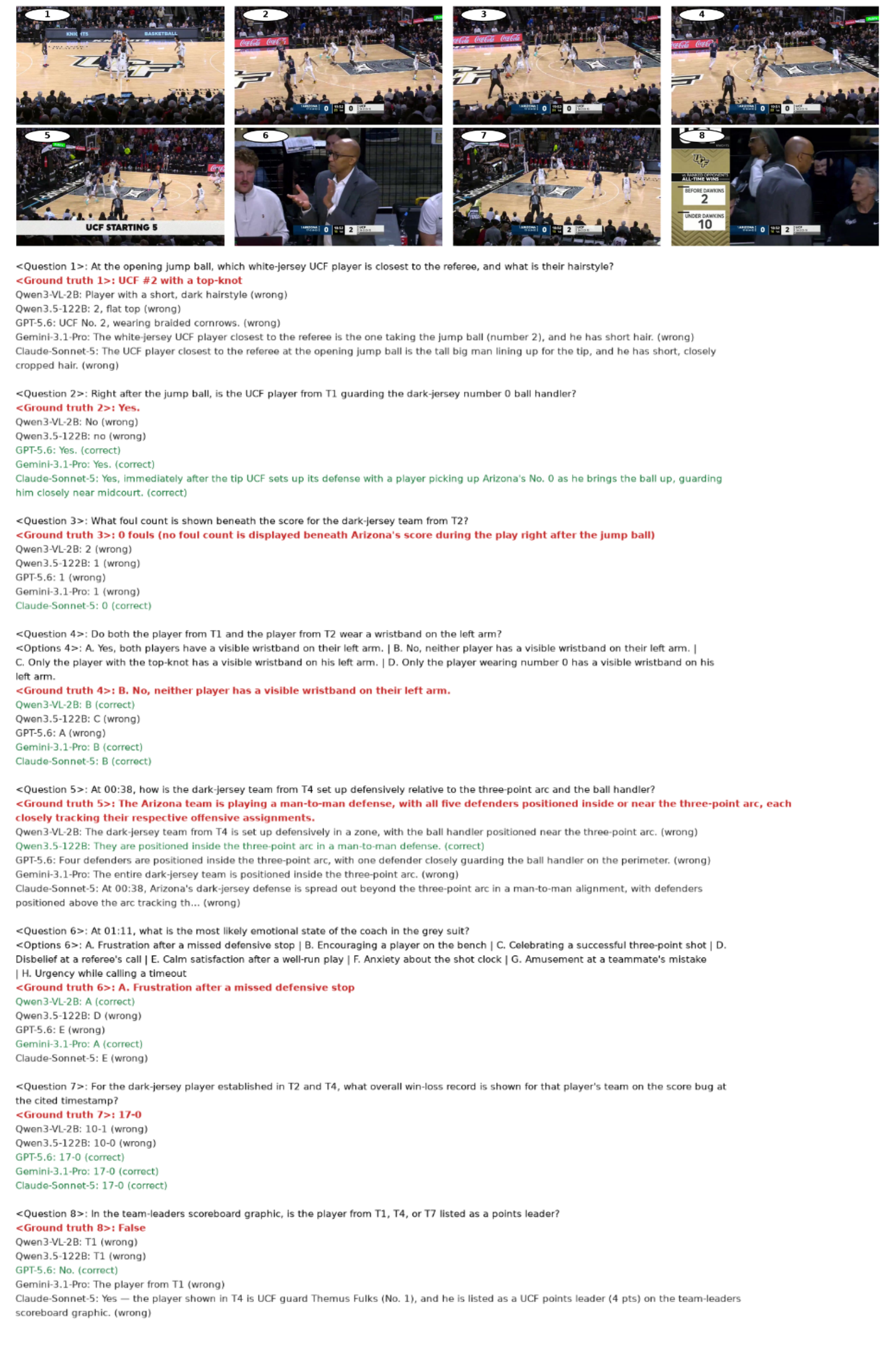}
\caption{\textbf{A basketball dialogue example from VideoMemBench.}
Selected video frames are shown above, followed by eight dialogue turns with reference answers and responses from five models. Reference answers are shown in red, and responses marked correct are shown in green.}
\label{fig:example_basketball}
\end{figure*}

\autoref{fig:example_soccer} and \autoref{fig:example_basketball} present more sports video dialogues in which later questions revisit previously discussed players and teams, illustrating the combined challenges of visual recognition, cross-turn reference resolution, and reasoning.

\end{document}